\documentclass[conference]{IEEEtran}
\IEEEoverridecommandlockouts

\usepackage{cite}
\usepackage[colorlinks=true,allcolors=blue]{hyperref}
\usepackage{graphicx}
\usepackage{amsmath,amssymb}
\usepackage{booktabs}
\usepackage{multirow}
\usepackage{tabularx}
\usepackage{url}
\usepackage{multirow}
\usepackage{array}
\usepackage{balance}
\usepackage{float}

\newcolumntype{Y}{>{\raggedright\arraybackslash}X}

\def\BibTeX{{\rm B\kern-.05em{\sc i\kern-.025em b}\kern-.08em
    T\kern-.1667em\lower.7ex\hbox{E}\kern-.125emX}}

\begin{document}

\title{Open-Vocabulary vs Supervised Learning Methods for Post-Disaster Visual Scene Understanding\\
\thanks{This work has received funding from the European Union’s Horizon Europe research and innovation programme under Grant Agreement No. 101168042, project TRIFFID (auTonomous Robotic aId For increasing First responders Efficiency). The views and opinions expressed in this paper are those of the authors only and do not necessarily reflect those of the European Union or the European Commission.}
}

\author{
\IEEEauthorblockN{
Anna Michailidou\IEEEauthorrefmark{1},
Georgios Angelidis\IEEEauthorrefmark{1},
Vasileios Argyriou\IEEEauthorrefmark{2},
Panagiotis Sarigiannidis\IEEEauthorrefmark{3},\\
Georgios Th. Papadopoulos\IEEEauthorrefmark{1}\IEEEauthorrefmark{4}
}

\IEEEauthorblockA{\IEEEauthorrefmark{1}Department of Informatics and Telematics, Harokopio University of Athens, Athens, Greece}
\IEEEauthorblockA{\IEEEauthorrefmark{2}Department of Networks and Digital Media, Kingston University, London, United Kingdom}
\IEEEauthorblockA{\IEEEauthorrefmark{3}Department of Electrical and Computer Engineering, University of Western Macedonia, Kozani, Greece}
\IEEEauthorblockA{\IEEEauthorrefmark{4}Archimedes, Athena Research Center, Athens, Greece}
\IEEEauthorblockA{Emails: \{amichailidou, gangelidis, g.th.papadopoulos\}@hua.gr, Vasileios.Argyriou@kingston.ac.uk, psarigiannidis@uowm.gr}
}

\maketitle
\begin{abstract}
Aerial imagery is critical for large-scale post-disaster damage assessment. Automated interpretation remains challenging due to clutter, visual variability, and strong cross-event domain shift, while supervised approaches still rely on costly, task-specific annotations with limited coverage across disaster types and regions. Recent open-vocabulary and foundation vision models offer an appealing alternative, by reducing dependence on fixed label sets and extensive task-specific annotations. Instead, they leverage large-scale pretraining and vision-language representations. These properties are particularly relevant for post-disaster domains, where visual concepts are ambiguous and data availability is constrained. In this work, we present a comparative evaluation of supervised learning and open-vocabulary vision models for post-disaster scene understanding, focusing on semantic segmentation and object detection across multiple datasets, including FloodNet+, RescueNet, DFire, and LADD. We examine performance trends, failure modes, and practical trade-offs between different learning paradigms, providing insight into their applicability for real-world disaster response. The most notable remark across all evaluated benchmarks is that supervised training remains the most reliable approach (i.e., when the label space is fixed and annotations are available), especially for small objects and fine boundary delineation in cluttered scenes.
\end{abstract}


\section{Introduction}

Rapid post-disaster situational awareness increasingly relies on remote sensing imagery from unmanned aerial vehicles (UAVs) and satellites, enabling wide-area mapping of hazards and damage \cite{cani2025triffid}. However, post-disaster aerial imagery is highly heterogeneous \cite{Rahnemoonfar23,Zhao24A,LADD,deVenancio2022}: targets exhibit extreme scale variation, scenes contain clutter, occlusions, and appearances shift substantially across different disasters and conditions. These factors, together with the limited availability of annotated data, continue to constrain reliable deployment and motivate robust learning strategies beyond narrow, fixed-vocabulary supervision.

Progress has been driven by (i) improved benchmarks for disaster and damage understanding in aerial and satellite imagery \cite{Rahnemoonfar21,Rahnemoonfar23,Gupta19xBD,Zhao24A} and (ii) advances in general-purpose computer vision architectures \cite{cani2026illicit,rodis2024multimodal} and pretraining \cite{konstantakos2025self,alimisis2025advances}. Convolutional backbones aided by large-scale supervised pretraining and architectural improvements \cite{Omar2024}, remain strong for dense prediction and real-time deployment, while vision transformers \cite{Dosovitskiy21} improved global context modeling for complex scenes. For object detection in particular, the field has evolved from fast one-stage detectors \cite{Redmon16} toward end-to-end set prediction with transformers \cite{Carion20}, including designs targeting real-time operation \cite{Zhao24B,lv2024rtdetrv2improvedbaselinebagoffreebies}.

A recent shift is the emergence of open-vocabulary (OV) and foundation models (FMs) that reduce dependence on fixed label sets, by leveraging vision--language pretraining \cite{Radford21} or promptable segmentation \cite{Kirillov23}. These models enable text-conditioned prediction in detection and segmentation and are attractive in disaster response, where target concepts may change and labeled data are limited.

In this work, aerial post-disaster perception is studied through a comparative evaluation that spans both semantic segmentation and object detection, contrasting closed-set supervised learning methods with open-vocabulary alternatives under a unified experimental protocol.
Our contributions include:
\begin{itemize}
  \item A comparative study of closed-set vs.open-vocabulary approaches for post-disaster aerial scene understanding, covering both semantic segmentation and object detection.
  \item An analysis of architectural trends across Convolutional Neural Networks (CNNs) and Transformers in cluttered aerial imagery.
  \item Practical insights into common failure modes (e.g., small objects, occlusions, visually ambiguous regions) and trade-offs relevant to operational deployment.
\end{itemize}
The remainder of this paper is organized as follows. Section~\ref{sec:related_work} reviews related work on supervised and open-vocabulary post-disaster scene understanding. Section~\ref{sec:methods} presents the methods selected for analysis, along with their taxonomy. Section~\ref{sec:experiments} describes the experimental protocol, and analyzes the corresponding results. Finally, Section~\ref{sec:conclusion} summarizes the conclusions and outlines potential directions for future work.

\section{Related Work}
\label{sec:related_work}

Public benchmarks have played a major role in standardizing evaluation practices and increasing the realism of post-disaster perception tasks, including flood scene understanding \cite{Rahnemoonfar21}, UAV-based damage segmentation \cite{Rahnemoonfar23}, and large-scale collections spanning multiple hazards \cite{Weber22}. Related satellite benchmarks for building damage assessment further illustrate the diversity of damage patterns and cross-event generalization challenges \cite{Gupta19xBD}. Yet, despite growing data resources, most studies remain scoped to a single hazard dataset, or supervision regime, reducing clarity regarding how architectural and learning choices transfer across disaster conditions.

In semantic segmentation, classic end-to-end dense prediction evolved from fully convolutional formulations \cite{Long15} and encoder--decoder designs with skip connections \cite{Ronneberger15} toward multi-scale context aggregation and boundary refinement, exemplified by DeepLab \cite{Chen18} atrous context and PSPNet \cite{Zhao17A} pyramid pooling. More recent work increased global reasoning via transformers \cite{Xie21} and masked-attention decoders \cite{Cheng22}, which can be beneficial for large homogeneous regions and complex spatial layouts typical of disaster imagery. In object detection, real-time one-stage detectors such as YOLO \cite{Redmon16}, remain widely used for aerial deployment, while transformer detectors \cite{Carion20} reframed detection as a set prediction without anchors or Non-Maximum Suppression (NMS) and made advances toward more robust training approaches and real-time variants \cite{Zhang22DINO,Zhao24B,Robinson25,Huang25}, often leveraging self-supervised representations \cite{Oquab23}.

More recently, OV and FMs have introduced language-conditioned prediction and promptable segmentation, reducing reliance on task-specific labels and enabling flexible, user-defined concepts at inference time \cite{Radford21,Kirillov23}. Open-vocabulary detection and grounding support text-conditioned localization \cite{Minderer22,Liu24B}, while open-vocabulary segmentation transfers vision--language semantics to the pixel space \cite{Dong23,Yu23,Xu23}. 


Post-disaster imagery poses atypical difficulties for these paradigms, such as heavy clutter, reflections, smoke, extreme scale variation, and cross-event domain shift. Existing studies largely evaluate such methods either outside disaster settings~\cite{ovseg} or within narrow single-dataset scopes~\cite{Gupta19xBD,Rahnemoonfar21, Rahnemoonfar23}. Most commonly, they do not include open-vocabulary methods~\cite{Asad23}, making it difficult to characterize failure modes and robustness under realistic disaster conditions.

In particular, there is, to the best of our knowledge, no head-to-head evaluation that (i) compares closed-set supervised pipelines against open-vocabulary alternatives under the same protocol, (ii) covers both semantic segmentation and object detection, and (iii) analyzes where and why open-vocabulary methods break down. This paper addresses these gaps by benchmarking representative supervised and open-vocabulary methods across both tasks on aerial post-disaster imagery, and by reporting performance trends and qualitative failure modes.

\section{Experimental Framework}
\label{sec:methods}

We evaluate two core tasks for post-disaster scene understanding across multi-hazard scenarios (earthquakes, floods, wildfires): (i) \emph{semantic segmentation}, assigning a label to every pixel, and (ii) \emph{object detection}, localizing targets of interest. Methods were selected to (a) reflect common baselines in disaster/remote-sensing literature and (b) cover complementary \emph{architectural} families (CNN, transformer) and \emph{learning} regimes (closed-set supervised vs.\ open-vocabulary/zero-shot). This supports controlled comparisons under clutter, scale variation, and domain shift.

\begin{figure}
    \centering
    \includegraphics[width=1.0\linewidth]{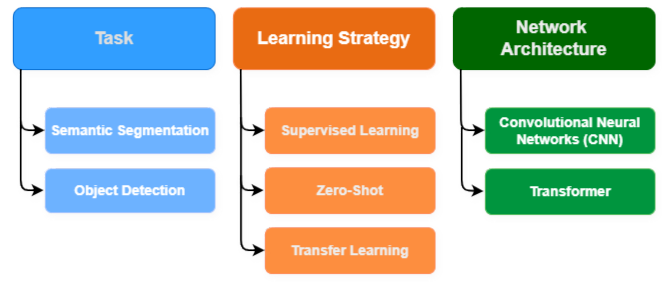}
    \caption{Taxonomy of post-disaster visual scene understanding methods}
    \label{fig:placeholder}
\end{figure}

\subsection{Closed-set Supervised Semantic Segmentation}
\label{subsec:methods_sup_seg}

For supervised segmentation, established CNN baselines and recent transformer designs are included to capture complementary inductive biases—multi-scale context aggregation and boundary refinement in CNNs versus global context modeling in transformers.

DeepLabV3+~\cite{Chen18} is a strong CNN baseline that combines atrous convolutions with Atrous Spatial Pyramid Pooling (ASPP) to capture multi-scale context, while maintaining good boundary quality in high-resolution scenes.

PSPNet~\cite{Zhao17A} aggregates global context through pyramid pooling, serving as a robust reference for aerial imagery where local textures can be ambiguous (e.g., shadows, reflections, debris).

CCNet~\cite{Huang19} introduces criss-cross attention to efficiently capture long-range dependencies with lower cost than full self-attention. We include it to test whether structured global-context attention improves performance in cluttered disaster scenes, relative to purely convolutional context aggregation.

SegFormer~\cite{Xie21} represents an efficient transformer segmentation design with a hierarchical Mix Transformer (MiT) encoder and lightweight decoder, offering long-range modeling, while remaining practical for large remote-sensing inputs.

Mask2Former~\cite{Cheng22} is included as a modern masked-attention segmentation framework. It is typically instantiated with strong backbones (e.g. ResNet~\cite{He16} or hierarchical transformers). We evaluate it strictly in the semantic setting, since our datasets provide semantic annotations.

\subsection{Open-Vocabulary Semantic Segmentation}
\label{subsec:methods_ov_seg}

To contrast closed-set supervision with approaches that reduce reliance on task-specific labels, OVS methods are evaluated under zero-shot text prompting using vision–language representations, and additionally a transfer-learning setting where the model is adapted to the target domain.

MaskCLIP~\cite{Dong23} adapts CLIP-style vision--language pretraining~\cite{Radford21} for dense prediction, typically using a CLIP ViT image encoder (ViT~\cite{Dosovitskiy21}) to transfer language-aligned semantics to pixel labels without training on target-dataset categories.

FC-CLIP~\cite{Yu23} performs open-vocabulary segmentation using \emph{frozen} CLIP-based features and a lightweight dense prediction head, enabling a strong assessment of how far language-aligned features transfer to high-variance post-disaster categories.

SegEarth-OV-3~\cite{Li2025} targets remote sensing Open-Vocabulary Semantic Segmentation (OVSS) under a training-free setup by adapting SAM3~\cite{sam3} to dense aerial scenes. It combines the semantic and instance output of SAM3 via a simple mask-fusion strategy and uses the presence scoring of the model to suppress absent categories, mitigating false positives from large vocabularies in patch-based RS processing. We treat it as a domain aware OV baseline for aerial disaster imagery.

\subsection{Closed-set Supervised Object Detection}
\label{subsec:methods_sup_det}

For object detection, real-time CNN detectors and efficient DETR-style transformers are included to compare CNN vs.\ transformer behavior under small-object and cluttered conditions.

YOLO series~\cite{Redmon16} covers the family of one-stage, real-time detectors employed here as a representative set of CNN baselines. YOLOv5 provides a widely adopted anchor-based reference with a CSPDarknet backbone and SPPF/PAN feature aggregation, while YOLOv8 and YOLO11 modernize the pipeline with anchor-free prediction and refined heads for improved accuracy–latency trade-offs. YOLO26~\cite{YOLO26} further targets edge deployment by removing DFL and enabling end-to-end, NMS-free inference to reduce post-processing overhead and latency jitter in real-time systems.

RT-DETRv2~\cite{lv2024rtdetrv2improvedbaselinebagoffreebies} is an improved real-time DETR-style detector that retains end-to-end set prediction while refining multi-scale feature handling for better practicality and robustness. It is included as a modern real-time transformer baseline whose selective multi-scale sampling and deployment-friendly operator choices aim to improve accuracy under clutter and scale variation without compromising speed.



\subsection{Open-Vocabulary Object Detection}
\label{subsec:methods_ov_det}

Open-vocabulary detection is evaluated using zero-shot text prompting, enabling localization of user-specified concepts without retraining on a fixed label set.

OWL-ViT~\cite{Minderer22} performs text-conditioned detection using a CLIP-aligned ViT image encoder (ViT~\cite{Dosovitskiy21}, CLIP~\cite{Radford21}), serving as a clean baseline for prompt-based localization.

Grounding DINO~\cite{Liu24B} combines DETR-style detection with grounded pretraining on image--text data, improving text-to-box grounding quality relative to earlier open-set detectors.

YOLOE~\cite{YOLOE2025} is a real-time open-vocabulary YOLO variant that extends one-stage detection/segmentation to text prompting, visual prompting, and prompt-free inference within a single model. It implements lightweight alignment modules for region–text and visual-prompt conditioning (RepRTA, SAVPE) and a prompt-free scheme (LRPC), Ultralytics’ ~\cite{ultralytics} prompt-free checkpoints ship with a fixed 4,585-class vocabulary (RAM++ tag set~\cite{RAMplusplus}), enabling “no-prompt” operation similar to standard closed-set YOLO at deployment.


\section{Experimental Evaluation}
\label{sec:experiments}

This section describes the experimental configuration and quantitative results for the selected semantic segmentation and object detection methods on disaster-focused remote sensing datasets covering earthquakes, floods, wildfires, and aerial human detection. All experiments follow a consistent protocol: dataset splits adhere to the respective dataset conventions where available, models are trained or fine-tuned using standard settings derived from reference implementations, and results are reported using widely adopted metrics for dense prediction and detection.

\subsection{Datasets}
\label{subsec:exp_datasets}

We evaluate semantic segmentation on RescueNet ~\cite{Rahnemoonfar23} for post-earthquake damage assessment, and FloodNet+ \cite{Zhao24A} for flood scene understanding. Object detection is evaluated on LADD \cite{LADD}, aerial pedestrian detection relevant for Search and Rescue (SAR) missions, and D-Fire \cite{deVenancio2022} for fire and smoke detection. These datasets expose complementary challenges, including severe clutter, scale variation, occlusions, and high intra-class variability.

\begin{table}[!t]
\centering
\caption{Datasets used in this work.}
\label{tab:datasets_compact}
\scriptsize
\setlength{\tabcolsep}{3.5pt}
\renewcommand{\arraystretch}{1.15}
\begin{tabular}{@{}l l l r r r@{}}
\toprule
\textbf{Dataset} & \textbf{Use-Case} & \textbf{Task} & \textbf{\#Imgs} & \textbf{\#Cls} & \textbf{Year} \\
\midrule
FloodNet+~\cite{Rahnemoonfar21} & Flood        & Sem.\ seg. & 2289  & 9  & 2024 \\
RescueNet~\cite{Rahnemoonfar23} & Earthquake   & Sem.\ seg. & 4494  & 10 & 2023 \\
D-Fire~\cite{deVenancio2022}    & Wildfire     & Detection  & 21527 & 2  & 2021 \\
LADD~\cite{LADD}                & SAR  & Detection  & 1365  & 1  & 2023 \\
\bottomrule
\end{tabular}
\end{table}

\begin{table*}[!t]
\centering
\caption{Overall segmentation performance on the RescueNet and FloodNet+ benchmarks.}
\label{tab:seg_rescue_flood}
\footnotesize
\setlength{\tabcolsep}{4pt}
\renewcommand{\arraystretch}{1.10}
\begin{tabular}{@{}l c c c c c@{}}
\toprule
\multicolumn{1}{c}{\multirow{2}{*}{\textbf{Model}}} &
\multirow{2}{*}{\textbf{Backbone}} &
\multirow{2}{*}{\textbf{Learning paradigm}} &
\multirow{2}{*}{\textbf{NN type}} &
\multicolumn{1}{c}{\textbf{RescueNet}} &
\multicolumn{1}{c}{\textbf{FloodNet+}} \\
\cmidrule(lr){5-5}\cmidrule(lr){6-6}
& & & & \textbf{mIoU} & \textbf{mIoU} \\
\midrule

\multicolumn{6}{@{}l@{}}{\textbf{Supervised Learning}} \\
PSPNet~\cite{Zhao17A}      & ResNet-101~\cite{He16} & \multirow{5}{*}{Supervised Learning} & \multirow{3}{*}{CNN} & 78.90\% & 76.87\% \\
CCNet~\cite{Huang19}       & ResNet-101~\cite{He16} &                                     &                      & \textbf{79.95}\% & 75.15\% \\
DeepLabV3+~\cite{Chen18}   & ResNet-101~\cite{He16} &                                     &                      & 78.49\% & 76.27\% \\
SegFormer~\cite{Xie21}     & MiT-B4~\cite{Xie21}    &                                     & \multirow{2}{*}{Transformer} & 76.10\% & 75.61\% \\
Mask2Former~\cite{Cheng22} & SwinT~\cite{Liu2021}   &                                     &                      & 77.55\% & \textbf{79.20}\% \\

\cmidrule(lr){1-6}

\multicolumn{6}{@{}l@{}}{\textbf{Open-vocabulary}} \\
MaskCLIP~\cite{Dong23}      & ViT~\cite{Dosovitskiy21}         & \multirow{3}{*}{Zero-Shot}         & \multirow{2}{*}{Transformer} & 11.21\% & 10.19\% \\
SegEarth-OV-3~\cite{Li2025} & SAM3~\cite{sam3}                 &                                   &                              & \textbf{26.16}\% & \textbf{46.08}\% \\
FC-CLIP~\cite{Yu23}         & ConvNeXt-L~\cite{Liu2022convnet} &                                   & CNN                          & 21.66\% & 45.43\% \\

\cmidrule(lr){1-6}

MaskCLIP~\cite{Dong23}      & ViT~\cite{Dosovitskiy21}         & \multirow{3}{*}{Transfer Learning} & \multirow{2}{*}{Transformer} & 28.94\% & 24.38\% \\
SegEarth-OV-3~\cite{Li2025} & SAM3~\cite{sam3}                 &                                   &                              & \textbf{52.63}\% & \textbf{53.92}\% \\
FC-CLIP~\cite{Yu23}         & ConvNeXt-L~\cite{Liu2022convnet} &                                   & CNN                          & 51.85\% & 52.34\% \\
\bottomrule
\end{tabular}
\end{table*}

\subsection{Evaluation Metrics}
\label{subsec:exp_metrics}

For semantic segmentation we report mean Intersection-over-Union (mIoU) over all $C$ classes, following the standard definition used in semantic segmentation benchmarks (e.g., PASCAL VOC \cite{Everingham10}). To expose class-specific behavior under strong pixel imbalance, we additionally report per-class IoU.

For object detection we report mean Average Precision at IoU threshold $0.50$ (mAP$_{50}$), consistent with common practice in detection benchmarks (e.g., PASCAL VOC) \cite{Everingham10}. Where supported by the evaluation protocol, we also report COCO-style mAP$_{[50:95]}$ to reflect localization quality across stricter IoU thresholds \cite{Lin14}.
\subsection{Experimental Setup}
\label{subsec:exp_setup}

Across experiments a unified training protocol is used to ensure fair comparison. Unless stated otherwise, the models are initialized from pretrained weights and optimized with AdamW optimiser with a learning rate of $0.01$ and weight decay $10^{-4}$ for up to 100 epochs with early stopping (patience 10), using batch size equal to $8$ and input resolution of $768$ pixels. For each experiment, standard data augmentation is applied to mitigate overfitting. All of the dataset splits follow the standard protocol used in each benchmark.

For transfer-learning, partial fine-tuning is performed, only updating a subset of deeper layers and task-specific heads during training. In this setting, AdamW is used with a learning rate of $10^{-4}$ and weight decay $0.05$ for 10 epochs, using batch size equal to $2$ and image resolution of $768$ pixels.

\subsection{Evaluation Results}
\label{subsec:exp_results}



\begin{table*}[!t]
\centering
\caption{Overall detection performance on the LADD and D-Fire benchmarks.}
\label{tab:det_ladd_dfire}
\footnotesize
\setlength{\tabcolsep}{4pt}
\renewcommand{\arraystretch}{1.10}
\begin{tabular}{@{}l c c c c c c c@{}}
\toprule
\multicolumn{1}{c}{\multirow{2}{*}{\textbf{Model}}} &
\multirow{2}{*}{\textbf{Backbone}} &
\multirow{2}{*}{\textbf{Learning Paradigm}} &
\multirow{2}{*}{\textbf{NN Type}} &
\multicolumn{2}{c}{\textbf{LADD}} &
\multicolumn{2}{c}{\textbf{D-Fire}} \\
\cmidrule(lr){5-6}\cmidrule(lr){7-8}
& & & & \textbf{mAP$_{50}$} & \textbf{mAP$_{50:95}$} & \textbf{mAP$_{50}$} & \textbf{mAP$_{50:95}$} \\
\midrule

\multicolumn{8}{@{}l@{}}{\textbf{Supervised Learning}} \\
YOLOv11l~\cite{Redmon16}  & CSP-based~\cite{ultralytics} & \multirow{3}{*}{Supervised Learning} & \multirow{3}{*}{CNN} & 92.2\% & 58.2\% & \textbf{80.6}\% & 48.9\% \\
YOLO26L~\cite{YOLO26}     & CSP-based~\cite{YOLO26}                   &                                      &                       & 93.6\% & 61.8\% & 80.3\% & \textbf{49.0}\% \\
RT-DETRv2-L~\cite{lv2024rtdetrv2improvedbaselinebagoffreebies} & ResNet~\cite{He16} &                                      &                       & \textbf{94.7}\% & \textbf{62.68}\% & 77.43\% & 43.97\% \\

\cmidrule(lr){1-8}

\multicolumn{8}{@{}l@{}}{\textbf{Open-vocabulary}} \\
YOLOE26~\cite{YOLOE2025}     & YOLO26-based~\cite{YOLO26}        & \multirow{3}{*}{Zero-Shot}         & CNN                          & 24.7\% & 11.3\% & 14.9\% & 7.1\% \\
Grounding DINO~\cite{Liu24B} & SwinT~\cite{Liu2021}              &                                   & \multirow{2}{*}{Transformer} & \textbf{61.0}\% & 28.2\% & 27.5\% & 13.4\% \\
OWL-ViT~\cite{Minderer22}    & ViT~\cite{Dosovitskiy21}          &                                   &                              & 6.2\% & 1.9\% & \textbf{36.4}\% & \textbf{18.0}\% \\

\cmidrule(lr){1-8}

YOLOE26~\cite{YOLOE2025}     & YOLO26-based~\cite{YOLO26}        & \multirow{3}{*}{Transfer Learning} & CNN                          & 37.5\% & 19.0\% & 24.2\% & 10.7\% \\
Grounding DINO~\cite{Liu24B} & SwinT~\cite{Liu2021}              &                                   & \multirow{2}{*}{Transformer} & \textbf{92.2}\% & \textbf{52.1}\% & \textbf{65.6}\% & \textbf{33.1}\% \\
OWL-ViT~\cite{Minderer22}    & ViT~\cite{Dosovitskiy21}          &                                   &                              & 68.6\% & 38.4\% & 48.9\% & 19.4\% \\
\bottomrule
\end{tabular}
\end{table*}

To emphasize the central focus of this study, results are organized by learning paradigm, closed-set supervised versus OV, and then discussed for each dataset.


\begin{figure*}[t]
    \centering
    \hspace*{0.9cm} 
    \includegraphics[width=0.9\linewidth]{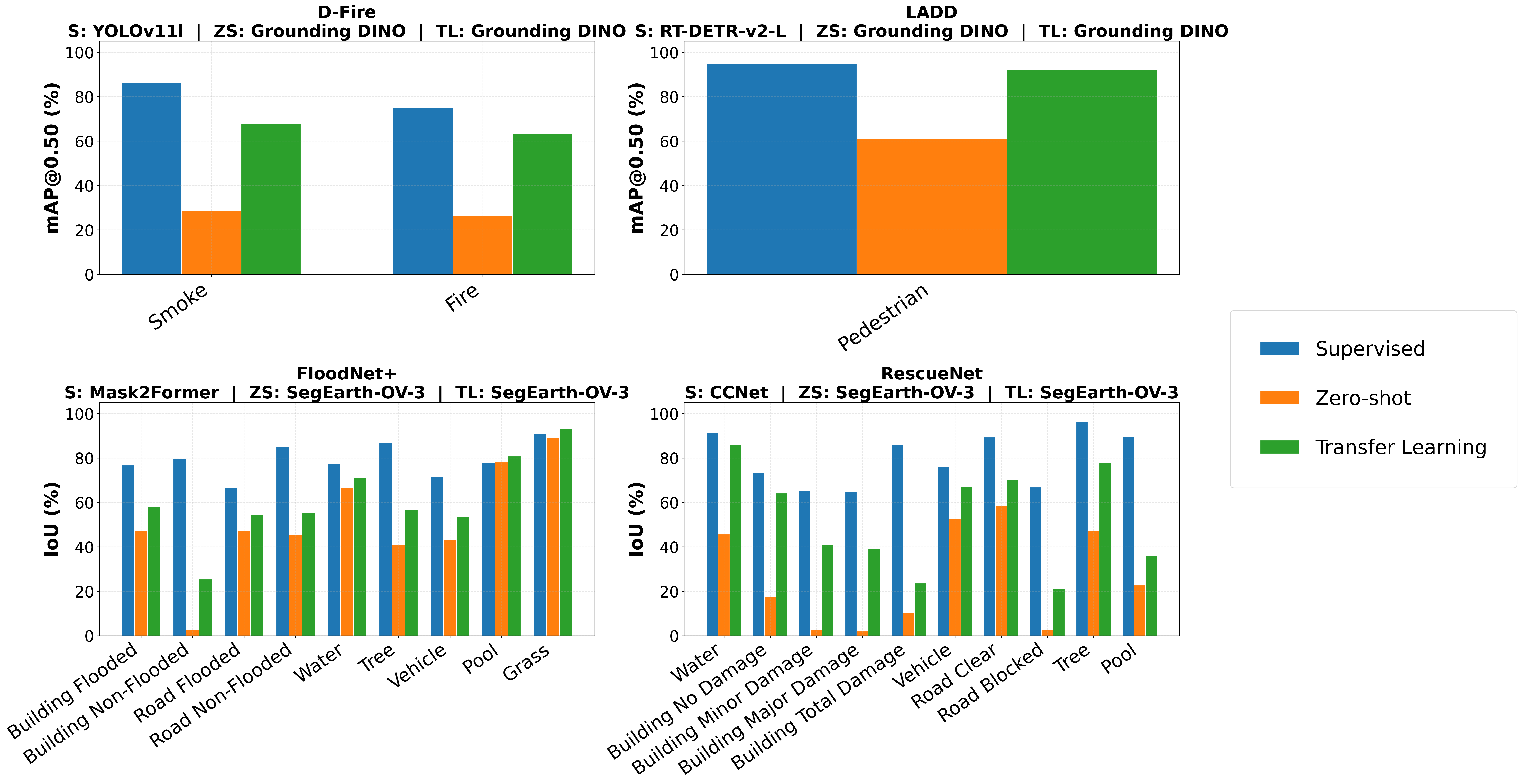}
    \caption{Per-class diagram of top performing models across different training setups for each dataset}
    \label{fig:perclass}
\end{figure*}

\subsubsection{Semantic Segmentation}
\label{subsec:exp_results_seg}

\paragraph{Closed-set supervised segmentation}
Across both segmentation datasets, supervised models substantially outperform OV methods as shown in Table \ref{tab:seg_rescue_flood}, consistent with direct optimization on the target label space and in-domain appearance statistics. In contrast, open-vocabulary pipelines must transfer semantics from large-scale pretraining into aerial post-disaster imagery with pronounced domain shift (damage textures, debris, reflections) and strong class imbalance.

Within supervised learning methods, differences are comparatively small and broadly align with architectural inductive biases. On RescueNet, CCNet slightly improves over PSPNet and DeepLabV3+, plausibly due to criss-cross attention propagating scene-level context, when damage evidence is spatially distributed. On FloodNet+, Mask2Former ranks highest, consistent with mask-level decoding that supports region reasoning and boundary refinement between large ``stuff'' areas, such as water and thin structures.

\paragraph{Open-vocabulary segmentation}
Zero-shot OV segmentation remains far below supervised performance reflecting limited alignment between vision--language pretraining and disaster-specific aerial scenes, as well as ambiguity in mapping dataset taxonomies to short prompts. After transfer learning, mIoU increases markedly, suggesting that even modest fine-tuning helps the models adjust their representations to the target domain. Performance gaps among OV methods persist due to mask quality under clutter and how well fine-tuning aligns predictions to the dataset taxonomy.

\subsubsection{Object Detection}
\label{subsec:exp_results_det}

\paragraph{Closed-set supervised detection}
Table~\ref{tab:det_ladd_dfire} shows a clear advantage for supervised detectors on both benchmarks, especially on LADD, where small aerial targets dominate. One-stage CNN detectors remain strong in small-object regimes due to dense multi-scale features and localization priors, while real-time DETR-style models benefit from global reasoning and end-to-end matching that can improve consistency under clutter.

\paragraph{Open-vocabulary detection: zero-shot vs.\ transfer learning}
Zero-shot OV detection is highly model- and dataset-dependent (Table~\ref{tab:det_ladd_dfire}), reflecting sensitivity of text--region grounding to domain shift and small-object visibility. Grounding-oriented designs generally provide stronger zero-shot localization than purely text-conditioned baselines, but aerial small objects and background clutter still induce under-detection and prompt sensitivity. Transfer learning reduces the gap on LADD, indicating that in-domain adaptation primarily improves proposal quality and localization, which then enables more effective text alignment.

\subsubsection{Dataset-specific observations}
\label{subsec:exp_results_datasetwise}

\paragraph{RescueNet}
Under closed-set supervision, visually distinctive classes (tree, water, road-clear) achieve consistently high IoU across architectures, indicating that in-domain training largely saturates these categories. The remaining errors concentrate on classes that require contextual interpretation, most notably the intermediate damage levels (minor/major) and road-blocked, where debris, shadows, and mixed materials reduce separability. CCNet attains the best overall mIoU and is comparatively stronger on these harder categories, consistent with the advantage of propagating scene-level context when local evidence is incomplete. In the open-vocabulary setting, zero-shot predictions mostly recover coarse semantics, while fine-grained damage levels and road-blocked remain weak, transfer learning substantially improves most classes, but the same context-heavy categories continue to limit performance.

\paragraph{FloodNet+}
For supervised segmentation, large “stuff” regions (grass, tree, road-non-flooded) are robustly segmented by all methods, whereas road-flooded and the smaller categories (vehicle, pool) remain the dominant sources of error. Mask2Former reaches the top mIoU largely by improving these difficult classes, suggesting better region-level refinement and boundary handling under reflective water and clutter. In open-vocabulary segmentation, zero-shot behavior is relatively stronger for broad natural categories than for flood-specific distinctions (flooded vs.\ non-flooded). Transfer learning produces the largest gains precisely on the flood-specific and small-object categories, indicating that limited adaptation primarily calibrates decision boundaries to the target taxonomy and imagery statistics.

\paragraph{LADD}
With a single target class (pedestrian), performance is governed by small-object recall and tight localization. The supervised results indicate that in-domain optimization is sufficient to achieve high sensitivity to small aerial instances under clutter, whereas open-vocabulary detectors are constrained by weak small-object grounding and only partially recover after transfer learning.

\paragraph{D-Fire}
Across supervised detectors, smoke is consistently easier than fire, consistent with smoke occupying larger regions, while fire is smaller, fragmented, and more variable in appearance. Open-vocabulary detection exhibits the same tendency in zero-shot mode and transfer learning improves both categories, by adapting proposal quality and localization to the aerial capture conditions.




\section{Conclusion}
\label{sec:conclusion}

This work presents a comparative evaluation of closed-set supervised and open-vocabulary methods for post-disaster aerial scene understanding, covering semantic segmentation and object detection across floods, earthquakes, wildfires and SAR. Across all benchmarks, supervised training remains the most reliable approach: when the label space is fixed and annotations are available, supervised models consistently achieve the highest accuracy and the most stable behavior, particularly for small targets and fine boundary delineation in cluttered scenes.

Open-vocabulary pipelines provide a practical alternative when dense labels are scarce, but zero-shot performance is markedly weaker under disaster-specific domain shift and fine-grained taxonomies. Transfer learning substantially reduces this gap, showing that open-vocabulary models are most effective as pretrained starting points rather than as drop-in replacements for supervised systems in high-resolution aerial imagery. Overall, the results indicate that the main challenges are shared across paradigms, small objects, occlusions, and subtle within-class variation, but supervision remains decisive for robust localization and boundary quality.

Future work will expand evaluation to broader hazards and acquisition conditions, improve open-vocabulary spatial precision for aerial scenes, and leverage foundation models for semi-automatic annotation to scale disaster-specific training data and strengthen cross-event generalization.

Moreover, the system will be evaluated with FRs to validate its effectiveness in simulated operational scenarios. Furthermore, the system will be connected with real-time AI-enabled event detectors \cite{linardakis2025survey, foteinos2025visual, linardakis2024distributed}, so as to facilitate scenarios involving Human-Robot Interaction \cite{papadopoulos2021towards, papadopoulos2022user, moutousi2025tornado} and broader security response incidents \cite{mademlis2024invisible}, while also supporting the necessary eXplainable Artificial Intelligence (XAI) pipelines \cite{rodis2024multimodal, evangelatos2025exploring}.

\balance

\bibliographystyle{IEEEtran}
\bibliography{bibliography.bib}

\end{document}